\documentclass[letterpaper]{article} 
\usepackage{aaai2027}  
\usepackage[hyphens]{url}  
\usepackage{graphicx} 
\usepackage{natbib}  
\usepackage{caption} 
\usepackage{algorithm}
\usepackage{algorithmic}

\usepackage{newfloat}
\usepackage{listings}
\DeclareCaptionStyle{ruled}{labelfont=normalfont,labelsep=colon,strut=off} 
\floatstyle{ruled}
\newfloat{listing}{tb}{lst}{}
\floatname{listing}{Listing}

\usepackage{booktabs}
\usepackage{amsmath,amssymb}

\title{EntropyMoE: Entropy-Aware Sparse Expert Routing for Tokenizer-Free LLMs}
\author {
    Bo Liu\textsuperscript{\rm 1},
    Muxuan Yu\textsuperscript{\rm 2},
    Yu Zhang\textsuperscript{\rm 3},
    Pengfei Gao\textsuperscript{\rm 2},
    Yongping Zhang\textsuperscript{\rm 2}
}
\affiliations {
    \textsuperscript{\rm 1}University of Bristol\\
    \textsuperscript{\rm 2}School of Automation Science and Electrical Engineering,Beihang University, Beijing, China\\
    \textsuperscript{\rm 3}University of Manchester\\
    sv25889@bristol.ac.uk, SY2403816@buaa.edu.cn, yu.zhang-43@postgrad.manchester.ac.uk, fpgao@buaa.edu.cn, yp\_zhang@buaa.edu.cn
}

\begin{document}

\maketitle

\begin{abstract}
Recent byte-level large language models (LLMs) have made tokenizer-free modeling increasingly competitive by grouping bytes into dynamically sized patches. However, existing byte-patch architectures still apply the same dense feed-forward computation to every patch. This uniform computation cannot adapt model capacity to variations in patch semantics and granularity. We address this limitation with EntropyMoE, a Mixture-of-Experts (MoE) architecture designed for dynamic byte patches.  EntropyMoE replaces the dense feed-forward modules in the global patch Transformer with Top-K expert layers. Each dynamic patch serves as the basic unit of expert routing, and its byte coverage determines its contribution to workload accounting. The router selects experts directly from patch entropy, using the same granularity signal that underlies dynamic patch construction to organize sparse computation. Patch entropy and length jointly define the feature space for regulating expert specialization. Experiments show that EntropyMoE achieves the lowest held-out bits-per-byte among matched dense and sparse baselines while maintaining comparable downstream accuracy. These results establish patch entropy as an effective routing coordinate for sparse conditional computation and extend Mixture-of-Experts modeling beyond tokenizer-based representations.
\end{abstract}

\section{Introduction}

Modern large language models (LLMs) typically rely on tokenization as the interface between raw text and neural computation. A tokenizer maps a byte or character stream to discrete subword units \citep{sennrich2016neural,kudo2018sentencepiece}, which an autoregressive Transformer processes to predict the next token \citep{vaswani2017attention}. Although effective, fixed-vocabulary units need not align with byte coverage, predictive uncertainty, or modeling difficulty. Suffixes, identifiers, multilingual byte sequences, and syntactically important symbols may be segmented into units with widely varying spans and uncertainty \citep{xue2022byt5,pagnoni2024blt}. Tokenizer quality also varies across languages and domains, affecting downstream performance \citep{rust2021tokenizer}. These limitations motivate interfaces that expose structure below fixed subword tokens.

Tokenizer-free LLMs address this limitation by operating directly on bytes \citep{xue2022byt5,yu2023megabyte,wang2024mambabyte}. In particular, the Byte Latent Transformer (BLT) makes byte-level modeling practical by grouping bytes into dynamically sized patches and performing the most expensive computation at the patch level \citep{pagnoni2024blt}. BLT determines patch boundaries from next-byte entropy: predictable regions are represented by longer patches, whereas uncertain regions are divided into shorter ones. Entropy therefore controls patch granularity while providing an explicit measure of local predictive uncertainty. Once the patches are formed, however, the global patch Transformer applies the same dense feed-forward transformation to every patch. The architecture adapts its segmentation to local uncertainty but does not use that signal to organize patch-level parameter activation.

Mixture-of-Experts (MoE) models offer a natural mechanism for conditional computation \citep{shazeer2017outrageously,lepikhin2021gshard,fedus2022switch}. By activating only a small subset of experts for each input unit, MoE architectures increase total parameter capacity while limiting the parameters used for any individual input. Existing MoE LLMs are nevertheless predominantly token-native: they route tokenizer-derived tokens, typically by projecting high-dimensional hidden states into expert logits \citep{fedus2022switch,jiang2024mixtral,dai2024deepseekmoe}. This formulation does not account for a setting in which the routed unit is a variable-length byte patch and an uncertainty signal has already been computed by the patching pipeline. It also leaves open whether a router must consume the full semantic representation when the model already exposes a causal, patch-level measure of local uncertainty. We therefore ask whether the entropy that determines dynamic byte patches can also serve as a compact coordinate for routing sparse expert computation.

We introduce \textbf{EntropyMoE}, a patch-native sparse expert architecture for tokenizer-free language modeling. Built on a BLT-style byte-patch pipeline, EntropyMoE replaces the dense feed-forward blocks in the global patch Transformer with Top-2 MoE layers and treats each dynamic patch as one routing unit. Unlike conventional hidden-state routers, its router maps only the scalar entropy associated with a patch to expert logits through a learned affine transformation. The patch hidden state is deliberately excluded from expert selection but remains the input to the selected feed-forward experts. Expert assignment is therefore organized by local uncertainty, while semantic and contextual information is preserved within expert computation.

\begin{figure*}[t]
  \centering
  \includegraphics[width=0.98\textwidth]{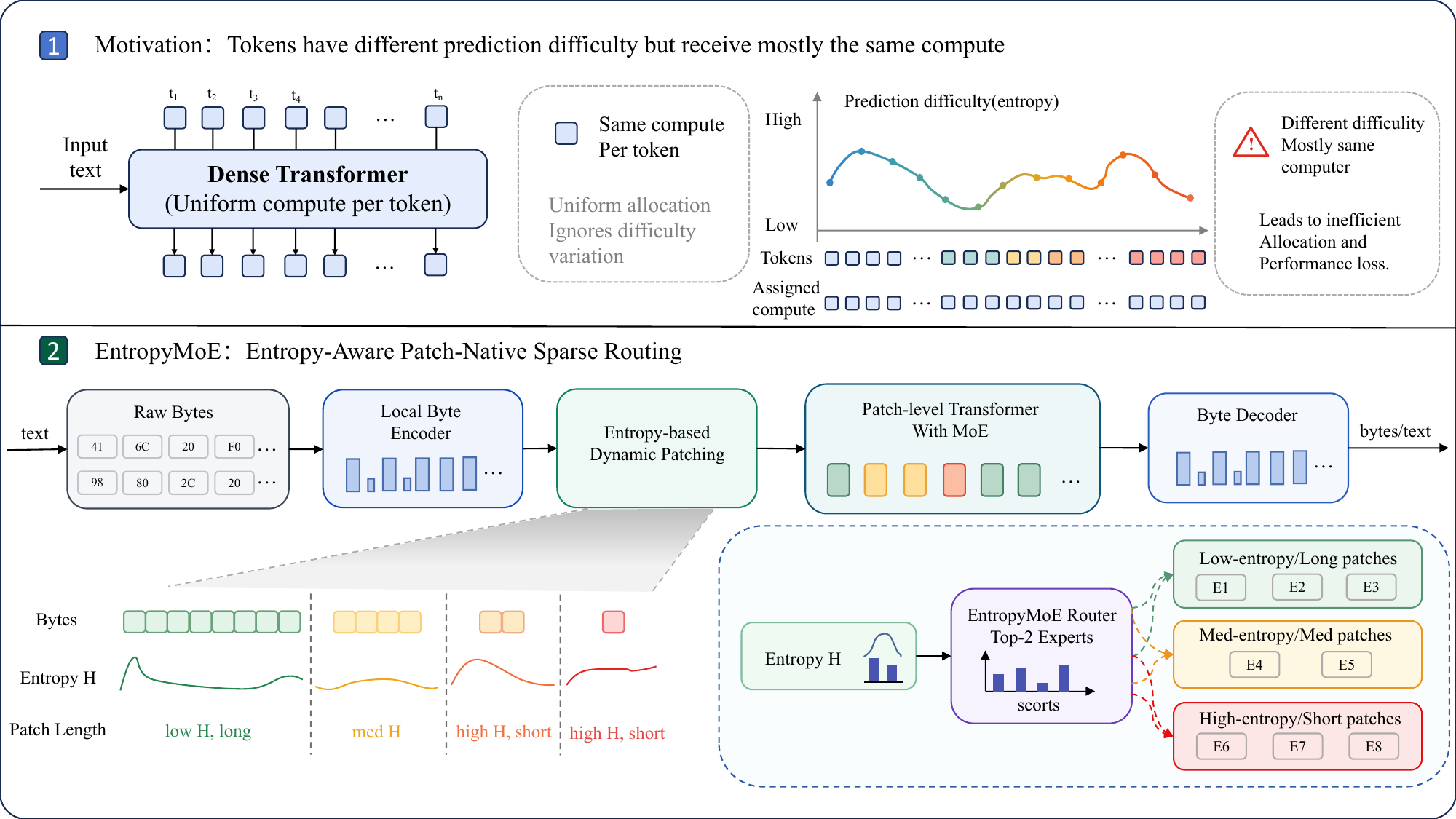}
  \caption{Overview of EntropyMoE. Raw bytes are grouped into entropy-based dynamic patches. The global patch Transformer replaces dense feed-forward blocks with Top-2 expert layers, and each patch is routed using scalar patch entropy.}
  \label{fig:entropymoe-overview}
\end{figure*}

EntropyMoE does not vary the number of active experts with patch difficulty: every valid patch activates exactly two experts. Adaptive granularity instead arises from the BLT patcher, which represents uncertain byte regions with more, shorter patches. The router determines which expert pair processes each patch, while patch length is used only to measure workload in represented bytes. Thus, patching determines computational granularity, scalar entropy determines the sparse parameter pathway, and byte length determines how routed workload is counted.

Across twenty-five layers with eight experts per layer, one slope and one bias per expert yield \(400\) effective router parameters, compared with more than \(4\times10^5\) for the hidden-state controls. EntropyMoE removes the hidden-state selection projection while retaining the same expert pool and Top-2 active capacity, testing whether a one-dimensional uncertainty signal can organize expert exposure.

Our controlled experiments compare EntropyMoE with Dense BLT, Hidden-only MoE, and Hidden+Entropy MoE under matched data, updates, active parameters, and dense-copy initialization. EntropyMoE achieves the lowest held-out bits per byte (BPB), whereas the hidden-state variants remain close to the dense baseline despite matching EntropyMoE's expert capacity. Its advantage over Hidden-only MoE is reproduced under a second continuation-training seed. Downstream accuracy remains broadly comparable, with a statistically supported improvement on HellaSwag rather than a universal gain across tasks. Sparse capacity alone therefore does not explain the observed modeling improvement.

Byte-weighted accounting consistently improves BPB over patch-count accounting across both tested seeds. EntropyMoE also separates low- and high-entropy assignment distributions more clearly than the hidden-state controls, although routing concentrates on a small expert subset. Its scalar router reduces routing parameters by three orders of magnitude and provides the best iteration time and valid-byte throughput among the sparse models. Nevertheless, dispatch, communication, and memory movement keep sparse execution slower than Dense BLT.

These findings support entropy-conditioned routing geometry under the evaluated continuation-training protocol, but do not establish indispensable expert identities or a compute-matched advantage over the dense backbone. We therefore present patch entropy as an effective low-dimensional routing coordinate while reserving stronger causal claims for future controlled studies.

Our main contributions are as follows:
\begin{enumerate}
\item We formulate patch-native sparse conditional computation for tokenizer-free LLMs, using dynamic byte patches as the fundamental units of expert routing.
\item We introduce EntropyMoE, an entropy-only Top-2 router with byte-coverage-weighted workload accounting, separating patch granularity, expert selection, and represented-byte load within a unified byte-patch architecture.
\item Through controlled dense and sparse comparisons, cross-seed continuation experiments, routing analysis, and implementation-cost auditing, we show that entropy routing improves byte-level modeling, induces structured entropy-conditioned assignments, and reduces routing overhead within the evaluated sparse family.
\end{enumerate}
\section{Related Work}
\paragraph{Dense tokenized LLMs.}
Dense tokenized LLMs remain the principal reference for modeling quality, scaling behavior, and training practice. The Transformer established self-attention as the standard sequence-modeling backbone, while BERT and GPT-style models demonstrated the effectiveness of large-scale pretraining \citep{vaswani2017attention,devlin2019bert,brown2020language}. Scaling-law and compute-optimal analyses then clarified the trade-offs among model capacity, data volume, and training compute \citep{kaplan2020scaling,hoffmann2022training}. Recent LLMs including PaLM, LLaMA/Llama, Qwen2.5, Gemma 3, and OLMo 2, have advanced this paradigm through improved data mixtures, longer contexts, architectural refinements, and increasingly transparent training recipes \citep{chowdhery2023palm,touvron2023llama,dubey2024llama,qwen2025qwen25,gemmateam2025gemma3,olmo2025olmo2}. These models provide useful reference points for tokenizer-free alternatives. Nevertheless, their computation remains organized around fixed subword tokens, causing token-level routing and load metrics to abstract away byte coverage and local predictive uncertainty. EntropyMoE addresses this interface mismatch by performing sparse expert routing over dynamically constructed byte patches.

\paragraph{Token-free and byte/patch-level LLMs.}
Token-free LLMs avoid a fixed subword vocabulary but must control the longer sequences induced by character or byte-level inputs. Character-level Transformers, CANINE, ByT5, and Charformer address this challenge through local processing, downsampling, or learned latent segmentation \citep{alrfou2019character,clark2022canine,xue2022byt5,tay2022charformer}. MEGABYTE introduces multiscale patch prediction, SpaceByte studies simple structure-aware patching, and MambaByte applies selective state-space modeling directly to bytes \citep{yu2023megabyte,slagle2024spacebyte,wang2024mambabyte}. BLT introduces entropy-based dynamic patches for global Transformer computation, while Fast BLT improves the efficiency of this architecture \citep{pagnoni2024blt,kallini2026fastblt}. Collectively, these works establish byte patches as practical latent units with variable span and modeling difficulty.These models nevertheless apply largely dense transformations at the latent level. EntropyMoE shifts the focus from forming efficient patches to using patch entropy for expert assignment and patch length for byte-aware workload accounting.

\paragraph{Mixture-of-experts and token-free conditional computation.}
MoE research has progressed from sparsely gated expert layers to token-routed Transformer systems such as GShard, Switch Transformer, GLaM, Mixtral, and DeepSeekMoE \citep{shazeer2017outrageously,lepikhin2021gshard,fedus2022switch,du2022glam,jiang2024mixtral,dai2024deepseekmoe}.Recent work has expanded this direction through transparent training platforms such as OLMoE and FLAME-MoE, together with large-scale systems such as DeepSeek-V3 and Qwen3 \citep{muennighoff2025olmoe,kang2025flamemoe,deepseekai2025deepseekv3,yang2025qwen3}. Recent routing and balancing studies further revisit the central MoE difficulty of assigning work to experts, including training-time routing dynamics, population-level load-balancing objectives, and differentiable expert allocation \citep{mouzouni2026threephases,chen2026phibalancing,zasada2026softmoe}. These methods are still primarily token-native: routers consume token hidden states and balance token counts or token-derived loads. EntropyMoE fills this interface gap by routing variable-length byte patches and defining expert load in byte, entropy, or cost-aware terms.

\section{Methodology}

We introduce \textbf{EntropyMoE}, a patch-native sparse expert architecture built on a BLT-style byte-patch model. It retains the local byte encoder and decoder, global self-attention, normalization, and residual structure, but replaces the global Transformer's dense feed-forward sublayers with Top-2 MoE layers. Each dynamic patch is one routed unit, and expert selection depends only on entropy produced by the patching pipeline. This design reuses a causal uncertainty signal and avoids a separate hidden-state routing projection.

\begin{figure*}[t]
  \centering
  \includegraphics[width=0.98\textwidth]{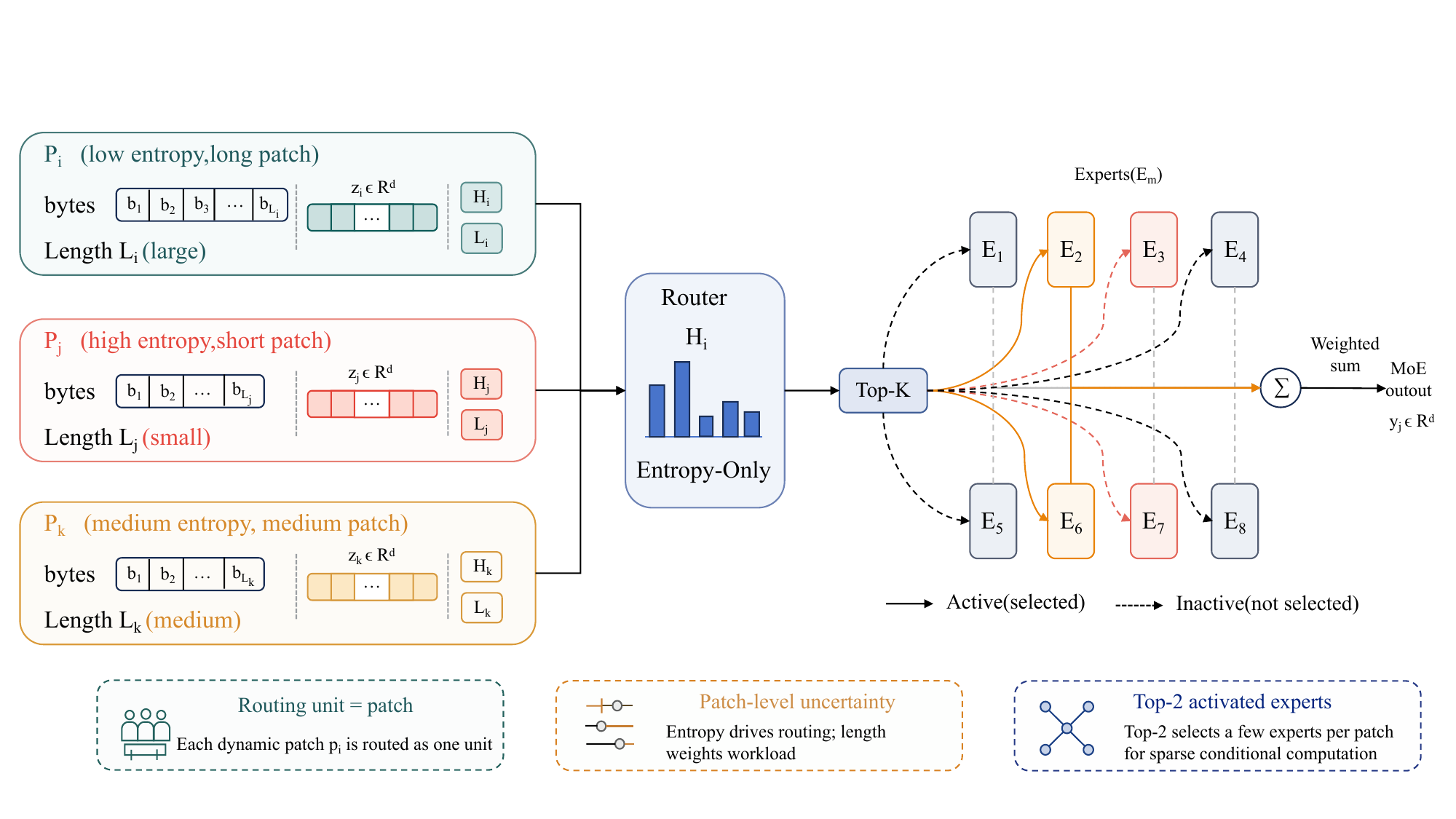}
  \caption{Patch-native entropy routing in EntropyMoE. Each dynamic byte patch is encoded into a latent representation for expert computation, while its normalized scalar entropy is the sole input to the Top-2 router. Patch length contributes only to byte-weighted workload accounting. The highlighted routing paths illustrate individual decisions rather than fixed expert identities across layers.}
  \label{fig:entropymoe-router}
\end{figure*}

\subsection{Patch-Native Conditional Computation}

Let \(\mathbf{x}_{1:T}=(x_1,\ldots,x_T)\) be a sequence of \(T\) UTF-8 bytes. Following the BLT byte-patch pipeline \citep{pagnoni2024blt}, a causal entropy model estimates next-byte uncertainty and an entropy-based patcher partitions the sequence into \(N\) contiguous dynamic patches:
\begin{equation}
P_i=\mathbf{x}_{b_i:b_{i+1}-1},
\qquad
i=1,\ldots,N.
\label{eq:dynamic-patches}
\end{equation}
Here \(\mathcal{P}=(P_1,\ldots,P_N)\) is the patch sequence, \(b_i\) and \(b_{i+1}\) are its boundaries, and \(L_i=b_{i+1}-b_i\) is the length of \(P_i\). Let \(H_i\) be the scalar patch entropy aggregated from causal next-byte entropy within \(P_i\). High-entropy regions tend to produce shorter patches, although \(H_i\) and \(L_i\) remain distinct.

A local byte encoder maps \(P_i\) to an initial representation \(\mathbf{u}_i^0\in\mathbb{R}^{d}\), and the global Transformer processes \((\mathbf{u}_1^0,\ldots,\mathbf{u}_N^0)\). Each patch is therefore one sparse routing unit. Before routing, \(H_i\) is normalized to \(\widetilde{H}_i\) to control feature scale without adding semantic or length features.

\subsection{EntropyMoE Architecture}

EntropyMoE replaces every global dense feed-forward sublayer with an expert pool while leaving the other backbone components unchanged. The main configuration contains twenty-five MoE layers, each with \(M=8\) experts and \(K=2\) active experts per valid patch. Each expert follows the replaced dense feed-forward structure. This conversion increases total capacity from \(1.46\)B to \(4.06\)B parameters while preserving \(1.46\)B active parameters per patch, isolating sparse routing from changes to byte encoding, attention, or decoding.

EntropyMoE does not allocate more experts to high-entropy patches. The patcher controls the number of patch units per byte span, whereas the router selects the expert pair for each unit. High-entropy regions receive more routing decisions per byte, but each retains the same Top-2 capacity.

\subsection{Scalar Patch-Entropy Routing}

At MoE layer \(\ell\), the router maps the normalized scalar entropy of patch \(P_i\) to an \(M\)-dimensional vector of expert logits:
\begin{equation}
\mathbf{s}_i^{\ell}
=
\mathbf{a}^{\ell}\widetilde{H}_i
+
\mathbf{b}^{\ell},
\qquad
\mathbf{a}^{\ell},\mathbf{b}^{\ell}\in\mathbb{R}^{M}.
\label{eq:entropy-router}
\end{equation}
Here \(\mathbf{a}^{\ell}\) and \(\mathbf{b}^{\ell}\) contain one learned entropy slope and one bias for each expert. The selected expert set is
\(\mathcal{S}_i^{\ell}=\operatorname{TopK}(\mathbf{s}_i^{\ell},K)\).
Because the input is scalar, each layer requires \(2M\) routing parameters, giving \(25\times2\times8=400\) effective parameters in total. A hidden-state router instead projects a \(d\)-dimensional patch state to \(M\) logits, requiring \(O(dM)\) rather than \(O(M)\) projection work per patch.

Layer-specific slopes and biases can partition the entropy axis differently across depth. Expert indices therefore have no fixed entropy regime across layers; within a layer, selection changes at intersections of the affine functions.

The patch state \(\mathbf{u}_i^{\ell}\) is processed by the selected experts but is excluded from expert selection. Hidden states, patch length, and byte-type features are not router inputs in the main model. Hidden-only and Hidden+Entropy MoE retain the same expert configuration and serve only as routing baselines.

\subsection{Sparse Expert Aggregation}

For patch \(P_i\), let \(\alpha_{i,e}^{\ell}\) be the mixture weight obtained by normalizing the selected logits over \(\mathcal{S}_i^{\ell}\), and let \(\mathbf{y}_i^{\ell}\) denote the sparse feed-forward output. The selected expert outputs are aggregated as
\begin{equation}
\mathbf{y}_i^{\ell}
=
\sum_{e\in\mathcal{S}_i^{\ell}}
\alpha_{i,e}^{\ell}
E_e^{\ell}(\mathbf{u}_i^{\ell}),
\qquad
\sum_{e\in\mathcal{S}_i^{\ell}}\alpha_{i,e}^{\ell}=1.
\label{eq:sparse-aggregation}
\end{equation}
Here \(E_e^{\ell}\) is expert \(e\) in layer \(\ell\). All patches activate the same number of experts, \(K=2\). EntropyMoE thus changes expert identity and mixture weight as a function of entropy, not the number of executed experts.

\subsection{Byte-Weighted Workload Accounting}

Dynamic patches represent different numbers of bytes, so patch counts do not measure how much byte-level data is associated with each expert. Let \(p_{i,e}^{\ell}=\operatorname{softmax}(\mathbf{s}_i^{\ell})_e\) be the soft probability before Top-2 truncation. EntropyMoE measures expert exposure using the byte-weighted routing mass
\begin{equation}
\rho_e^{\ell}
=
\frac{\sum_i L_i p_{i,e}^{\ell}}
{\sum_i L_i}.
\label{eq:byte-weighted-mass}
\end{equation}
An auxiliary term penalizes deviations of \(\rho_e^{\ell}\) from \(1/M\), is averaged across layers, and is added to the language-modeling loss with coefficient \(\lambda_{\mathrm{bal}}=0.02\). Patch length appears only in this objective and does not change the logits or Top-2 set. The objective measures soft exposure in bytes, whereas execution remains one Top-2 dispatch per patch.

\subsection{Dense-to-Sparse Initialization and Controlled Comparison}

All sparse models start from the same dense BLT checkpoint. Each dense feed-forward block is copied to every expert, while the router is initialized separately. Because the copied experts initially implement the same function and their normalized weights sum to one, each sparse layer initially reproduces the dense transformation. The experts diverge during continuation training as they receive different assignments.

EntropyMoE, Hidden-only MoE, and Hidden+Entropy MoE share this conversion, expert configuration, parameter counts, data, update budget, and warm start; they differ only in router inputs. GPU-hours are reported separately because implementation costs are not matched.

EntropyMoE consequently separates the main functions of the byte-patch MoE pipeline: entropy-based patching determines patch boundaries and computational granularity, normalized patch entropy determines expert assignment, patch length defines byte-aware auxiliary workload, and the evolving hidden representation supplies the semantic content transformed by the selected experts.

\section{Experiments}
\label{sec:experiments}

We evaluate EntropyMoE from four perspectives. Under matched data,
optimization, warm start, and activated capacity settings, we first compare
its modeling performance with dense and sparse routing baselines. We then
examine the robustness of the improvements and their transfer to downstream
tasks. Finally, we ablate byte-weighted load balancing, analyze the expert
routing patterns induced by entropy, and quantify the training overhead of
each implementation.

\subsection{Experimental Setup}
\label{sec:exp-setup}

\paragraph{Models.}
All controlled experiments use BLT-1B \citep{pagnoni2024blt} as the backbone.
The dense BLT baseline contains 1.46B parameters, all of which are activated
for each patch. For each sparse variant, we replace the 25 feed-forward blocks
in the global Transformer with MoE layers comprising eight experts, of which
two are activated per patch. This configuration contains 4.06B total
parameters and activates 1.46B parameters per patch. The sparse models differ only in their routing inputs.
EntropyMoE routes patches using scalar patch entropy, Hidden-only MoE uses the
patch hidden state, and Hidden+Entropy MoE augments the hidden-state router
with an entropy residual. All sparse variants use the same dense-copy
initialization, allowing their comparison to isolate the effect of router
input from differences in pretraining history.

\paragraph{Training.}
All runs use seed 42 and the same ordered subset of FineWeb-Edu \citep{penedo2024fineweb}. We train each model for 30,000
updates using AdamW with a learning rate of $10^{-5}$, weight decay of $0.1$,
and a constant learning-rate schedule. Training uses BF16 precision, a
sequence length of 256, and a global batch size of six distributed across six
A100 80GB GPUs. For the sparse models, the byte-weighted load-balancing
coefficient is set to $0.02$ unless otherwise stated. The primary quality
comparison controls for training data, update count, activated parameter
count, and sparse-model initialization. We report GPU-hours separately to
characterize realized implementation cost, but do not treat them as a matching
criterion.

\begin{table}[!t]
\centering
\scriptsize
\setlength{\tabcolsep}{2.0pt}
\caption{Controlled seed-42 quality results. All models use the same data,
updates, and active parameter count. Sparse models also share the same
dense-copy warm start. Lower BPB is better; downstream metrics are accuracy
(\%). Avg-6 is the unweighted mean over the six non-MMLU tasks.}
\label{tab:controlled-quality}
\textit{(a) Aggregate quality and model capacity.}\\[-2pt]
\begin{tabular*}{\columnwidth}
{@{\extracolsep{\fill}}@{}llrrr@{}}
\toprule
Model
& Total / active
& BPB$\downarrow$
& Avg-6$\uparrow$
& MMLU \\
\midrule
Dense BLT
& 1.46B / 1.46B
& 0.8442
& 50.20
& 24.73 \\
Hidden-only MoE
& 4.06B / 1.46B
& 0.8432
& 50.18
& 24.31 \\
Hidden+Entropy MoE
& 4.06B / 1.46B
& 0.8430
& 49.95
& 24.34 \\
EntropyMoE
& 4.06B / 1.46B
& \textbf{0.8351}
& \textbf{50.31}
& \textbf{25.10} \\
\bottomrule
\end{tabular*}

\vspace{3pt}

\textit{(b) Task-level accuracy (\%).}\\[-2pt]
\begin{tabular*}{\columnwidth}
{@{\extracolsep{\fill}}@{}lrrrrrr@{}}
\toprule
Model & PIQA & HellaSwag & ARC-E & ARC-C & OBQA & BoolQ \\
\midrule
Dense BLT
& 72.09 & 55.61 & 57.50 & \textbf{37.29}
& 26.40 & \textbf{52.32} \\
Hidden-only MoE
& 71.93 & 55.84 & \textbf{58.55} & 36.61
& 26.60 & 51.53 \\
Hidden+Entropy MoE
& 71.87 & 55.56 & 56.97 & 36.27
& \textbf{27.00} & 52.05 \\
EntropyMoE
& \textbf{72.42} & \textbf{56.76} & 57.14 & 36.61
& 26.80 & 52.14 \\
\bottomrule
\end{tabular*}
\end{table}

\paragraph{Evaluation.}
We measure held-out language-modeling quality using bits per byte (BPB). The
primary evaluation uses a fixed stream of 256 batches containing 1,833,696
valid target bytes, with identical batch boundaries and per-batch byte counts
across models. For downstream evaluation, answer choices are ranked by their
model likelihoods. We evaluate on PIQA, HellaSwag, ARC-Easy, ARC-Challenge,
OpenBookQA, BoolQ, and MMLU. The workload-accounting ablation uses a separately
constructed but internally matched stream of 256 batches containing 1,834,272
valid target bytes.

\paragraph{Uncertainty.}
For BPB, we construct 95\% confidence intervals using 10,000 paired bootstrap
replicates. For PIQA and HellaSwag, we additionally report paired confidence
intervals and exact two-sided McNemar tests. We apply Holm correction across
the six reported hypothesis tests. For the workload-accounting ablation, we
use 50,000 paired bootstrap replicates.

\subsection{Language Modeling Results}
\label{sec:main-results}

\paragraph{Overall comparison.}
EntropyMoE obtains the lowest BPB among all four controlled configurations
(Table~\ref{tab:controlled-quality}). Its paired improvement over every
baseline is larger than 0.0079 BPB, and every interval lies strictly below
zero (Table~\ref{tab:bpb-evidence}(a)). The gain is not explained by sparse
capacity alone: Hidden-only MoE and Hidden+Entropy MoE remain close to Dense BLT
despite matching EntropyMoE's total and active parameter counts.

\begin{table}[!t]
\centering
\scriptsize
\setlength{\tabcolsep}{2.0pt}
\caption{Statistical support for the BPB result. In (a), $\Delta$ is
EntropyMoE minus the baseline, so negative values favor EntropyMoE. In (b),
difference is Hidden-only minus EntropyMoE.}
\label{tab:bpb-evidence}

\textit{(a) Paired seed-42 comparisons.}\\[-2pt]
\begin{tabular*}{\columnwidth}
{@{\extracolsep{\fill}}@{}lcc@{}}
\toprule
Baseline & $\Delta$ BPB & Paired-bootstrap 95\% CI \\
\midrule
Dense BLT
& $-0.009152$
& $[-0.009732,-0.008575]$ \\
Hidden-only MoE
& $-0.008137$
& $[-0.008785,-0.007514]$ \\
Hidden+Entropy MoE
& $-0.007970$
& $[-0.008575,-0.007370]$ \\
\bottomrule
\end{tabular*}

\vspace{3pt}

\textit{(b) Matched within-seed replication.}\\[-2pt]
\begin{tabular*}{\columnwidth}
{@{\extracolsep{\fill}}@{}lccc@{}}
\toprule
Seed & EntropyMoE & Hidden-only & Difference \\
\midrule
42 & 0.835065 & 0.843203 & $+0.008137$ \\
43 & 0.835101 & 0.842379 & $+0.007278$ \\
\bottomrule
\end{tabular*}
\end{table}

\paragraph{Seed replication.}
The EntropyMoE consistently outperforms Hidden-only MoE across both continuation-training seeds, with comparable gains in each case (Table~\ref{tab:bpb-evidence}(b)). This consistency supports robustness to continuation-training randomness, while variability across independent pretraining runs remains unmeasured.

\subsection{Downstream Evaluation}
\label{sec:downstream}
EntropyMoE preserves competitive performance across the downstream suite while
achieving the highest Avg-6 and MMLU scores
(Table~\ref{tab:controlled-quality}). Avg-6 differs by at most 0.36 points
across the controlled models, indicating that the BPB improvement does not
compromise overall downstream quality. On HellaSwag, EntropyMoE outperforms
Dense BLT, Hidden-only MoE, and Hidden+Entropy MoE by 1.16, 0.93, and 1.20
percentage points, respectively.The improvements remain significant after Holm correction
(Table~\ref{tab:downstream-paired}). PIQA performance remains comparable across
models. These results establish BPB as the primary quality gain, complemented
by a statistically robust improvement on HellaSwag and competitive transfer
performance across the broader evaluation suite.

\begin{table}[!t]
\centering
\small
\setlength{\tabcolsep}{2.7pt}
\caption{Paired downstream comparisons. $\Delta$ is EntropyMoE accuracy minus
baseline accuracy in percentage points. E-only/B-only gives discordant
correctness counts.}
\label{tab:downstream-paired}
\resizebox{\columnwidth}{!}{%
\begin{tabular}{llrrrr}
\toprule
Task & Baseline & $\Delta$ & 95\% CI & E-only/B-only & $p_{\mathrm{Holm}}$ \\
\midrule
PIQA & Dense BLT & $+0.33$ & $[-0.65,+1.31]$ & 43/37 & 0.9609 \\
PIQA & Hidden-only MoE & $+0.49$ & $[-0.49,+1.47]$ & 47/38 & 0.9609 \\
PIQA & hidden+entropy & $+0.54$ & $[-0.44,+1.52]$ & 46/36 & 0.9609 \\
\midrule
HellaSwag & Dense BLT & $+1.16$ & $[+0.75,+1.57]$ & 288/172
& $4.24\!\times\!10^{-7}$ \\
HellaSwag & Hidden-only MoE & $+0.93$ & $[+0.51,+1.34]$ & 275/182
& $6.31\!\times\!10^{-5}$ \\
HellaSwag & hidden+entropy & $+1.20$ & $[+0.78,+1.63]$ & 311/190
& $4.24\!\times\!10^{-7}$ \\
\bottomrule
\end{tabular}}
\end{table}

\subsection{Workload-Accounting Ablation}
\label{sec:byte-accounting}

\paragraph{Balancing unit.}
Byte-weighted load balancing consistently improves BPB across both continuation-training seeds, with a mean reduction of $5.26\times10^{-4}$ BPB
(Table~\ref{tab:gateb-two-seed}). These repeatable gains support byte coverage
as an effective workload measure for balancing computation across
variable-length patches.

\begin{table}[!t]
\centering
\small
\setlength{\tabcolsep}{3.0pt}
\caption{Byte-weighted versus patch-count workload accounting. $\Delta$ is
patch-count BPB minus byte-weighted BPB. Intervals are paired batch-bootstrap
95\% intervals within each continuation-training seed.}
\label{tab:gateb-two-seed}
\resizebox{\columnwidth}{!}{%
\begin{tabular}{lrrrr}
\toprule
Seed
& Byte weighted
& Patch count
& $\Delta$
& 95\% CI
\\
\midrule
42 & \textbf{0.851015} & 0.851638 & $+0.000623$
& $[+0.000325,+0.000930]$ \\
43 & \textbf{0.851373} & 0.851801 & $+0.000429$
& $[+0.000121,+0.000736]$ \\
\bottomrule
\end{tabular}}
\end{table}

\begin{figure*}[!t]
  \centering
  \includegraphics[width=0.49\textwidth]
  {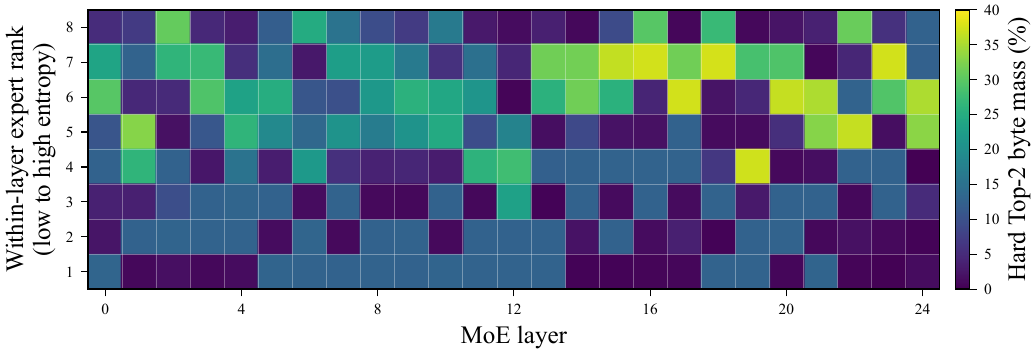}\hfill
  \includegraphics[width=0.49\textwidth]
  {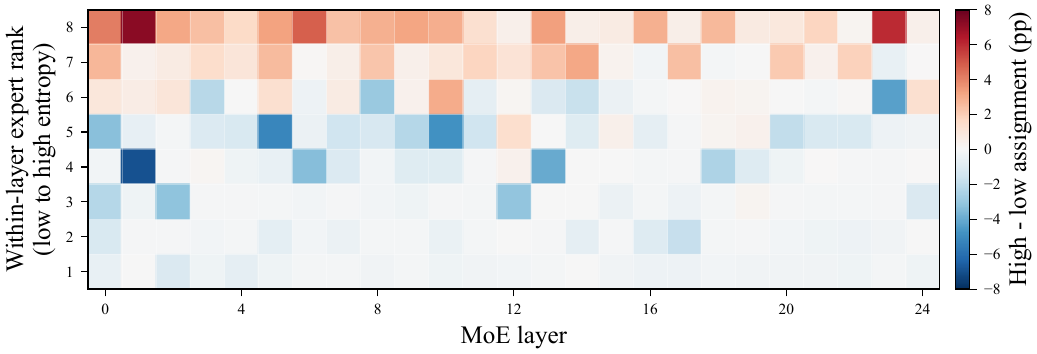}
  \caption{Layerwise routing in EntropyMoE. Experts are ranked independently
  within each layer by the mean entropy of their assigned patches (rank 1 is
  lowest and rank 8 is highest). Left: byte-weighted hard Top-2 assignment
  mass. Right: the difference between high- and low-entropy hard-assignment
  rates; red favors high-entropy patches and blue favors low-entropy patches.
  Expert ranks do not denote shared identities across layers.}
  \label{fig:layerwise-routing}
\end{figure*}

\paragraph{Accounting validation.}
Across all layers and batches, the logged byte-load distribution satisfies
\[
q^{\mathrm{byte}}_e
=
\operatorname{Normalize}\!\left(
q^{\mathrm{patch}}_e\,\overline{\ell}_e
\right),
\]
where $q^{\mathrm{patch}}_e$ denotes the fraction of patches assigned to expert
$e$, and $\overline{\ell}_e$ denotes their mean length. The reconstructed load
fractions agree with the logged values to within $3.4\times10^{-8}$, confirming
that byte coverage is accurately reflected in workload accounting. Together
with the consistent BPB improvement across both continuation-training seeds,
this result supports byte weighting as a principled accounting unit for
variable-length patches. Layer-averaged concentration varies across seeds,
suggesting that its primary benefit lies in accurately measuring expert
workload rather than uniformly flattening hard routing assignments.

\subsection{Routing Analysis}
\label{sec:routing-geometry}

EntropyMoE learns a markedly more selective and entropy-responsive routing
structure. Hidden-only MoE distributes assignments broadly and produces similar
routing profiles for low- and high-entropy patches. In contrast, EntropyMoE
directs most assignments to a compact expert subset: averaged across layers,
3.12 experts receive at least 1\% of the assignment mass, and the two most active
experts account for 89.1\%. It also exhibits substantially greater divergence
between low- and high-entropy routing distributions than the other sparse
variants. These results indicate that patch entropy provides a strong signal for
differentiating expert utilization across patch regimes.

\begin{table}[h]
\centering
\small
\setlength{\tabcolsep}{3.6pt}
\caption{Routing geometry averaged over 25 MoE layers. Active experts have at
least 1\% assignment mass; $N_{\mathrm{eff}}$ is the effective expert count;
Top-2 mass is the assignment mass of the two most used experts; entropy JSD
compares low- and high-entropy assignment distributions.}
\label{tab:mechanism-results}
\resizebox{\columnwidth}{!}{%
\begin{tabular}{lrrrr}
\toprule
Model
& Active exp.
& $N_{\mathrm{eff}}$
& Top-2 mass
& Entropy JSD
\\
\midrule
Hidden-only MoE & 7.84 & 5.94 & 0.554 & 0.0040 \\
Hidden+Entropy MoE  & 5.04 & 3.70 & 0.767 & 0.0024 \\
EntropyMoE      & 3.12 & 2.59 & 0.891 & \textbf{0.1263} \\
\bottomrule
\end{tabular}}
\end{table}

\paragraph{Layerwise load and entropy shift.}
Figure~\ref{fig:layerwise-routing} reveals the layerwise structure underlying
the aggregate concentration in Table~\ref{tab:mechanism-results}. A compact
subset of expert ranks carries most of the byte-weighted load, while the
dominant rank varies across depth. The assignment-rate differences show a
systematic entropy-dependent shift within each layer: high-entropy patches
preferentially select the upper entropy-ranked experts, whereas low-entropy
patches are more strongly associated with several middle ranks. The large
entropy JSD therefore reflects clear changes in hard expert assignments, not
merely differences in soft routing probabilities. Together, these results
demonstrate a depth-dependent, entropy-conditioned routing geometry that does
not rely on fixed expert identities across layers.

\subsection{Training Efficiency}
\label{sec:training-efficiency}

Table~\ref{tab:realized-cost} shows that all runs complete without OOM events
or allocation retries. Among the sparse models, EntropyMoE achieves the most
efficient training profile. Its scalar router uses $1024\times$ fewer trainable
parameters than the hidden-state router, reduces median iteration time by
8.6\%, and increases valid-byte throughput by 9.5\%. These results demonstrate
that entropy-based routing translates its compact parameterization into
measurable training-efficiency gains. The current sparse implementation remains
approximately twice as slow as Dense BLT, indicating substantial scope for
further improvement through optimized MoE kernels. Router counts exclude
dormant compatibility tensors, and GPU-hours report realized implementation
cost separately from the controlled quality comparison.

\begin{table}[h]
\centering
\scriptsize
\setlength{\tabcolsep}{2.0pt}
\caption{Implementation and realized-cost audit. Iteration time and valid-byte
throughput are steady medians over steps 1,001--29,000; memory values
are observed peaks. GPU-hours are elapsed job time multiplied by six GPUs.}
\label{tab:realized-cost}
\textit{(a) Router and training cost.}\\[-2pt]
\resizebox{\columnwidth}{!}{%
\begin{tabular}{lrrrr}
\toprule
Model
& Router params
& Iter. time (s)
& Valid bytes/s
& GPU-hours
\\
\midrule
Dense BLT
& -- & 0.6433 & 11,156.7 & 33.81 \\
Hidden-only MoE
& 409,600 & 1.4084 & 5,080.9 & 73.40 \\
Hidden+Entropy MoE
& 410,000 & 1.3390 & 5,351.0 & 68.69 \\
EntropyMoE
& 400 & 1.2871 & 5,561.2 & 66.16 \\
\bottomrule
\end{tabular}}
\vspace{3pt}

\textit{(b) Memory and dense-relative cost.}\\[-2pt]
\resizebox{\columnwidth}{!}{%
\begin{tabular}{lrrrr}
\toprule
Model
& Active GiB
& Reserved GiB
& Time / Dense
& Throughput / Dense
\\
\midrule
Dense BLT
& 22.14 & 30.18 & 1.000$\times$ & 1.000$\times$ \\
Hidden-only MoE
& 38.15 & 55.50 & 2.189$\times$ & 0.455$\times$ \\
Hidden+Entropy MoE
& 38.10 & 48.87 & 2.081$\times$ & 0.480$\times$ \\
EntropyMoE
& 38.10 & 50.29 & 2.001$\times$ & 0.498$\times$ \\
\bottomrule
\end{tabular}}
\end{table}

Overall, EntropyMoE is the most efficient sparse router tested, while the
current sparse implementation remains more expensive than Dense BLT. The
efficiency result therefore supports the practical benefit of scalar routing
within the sparse family; it does not claim a lower end-to-end training cost
than the dense backbone.

\section{Conclusion}
\label{sec:conclusion}

We introduced EntropyMoE, a patch-native sparse architecture that routes dynamic byte patches to Top-2 experts using only scalar causal patch entropy. By separating expert selection from high-dimensional hidden representations, EntropyMoE reduces the routing network to 400 trainable parameters across 25 MoE layers, while preserving semantic and contextual information within expert computation. In controlled BLT-1B continuation experiments, EntropyMoE achieved the lowest held-out BPB among the dense and sparse configurations tested under matched data, updates and active parameter count. Its advantage over hidden-state routing was consistent across two continuation seeds, while downstream accuracy remained broadly comparable. EntropyMoE also achieved the lowest routing overhead among the evaluated sparse models, although the current sparse implementation remained more expensive than Dense BLT.

Our analysis further showed that scalar entropy induces distinct, albeit concentrated, expert-assignment patterns across patch-entropy regimes. These patterns demonstrate entropy-conditioned routing geometry, but they have not yet established functional expert specialization. Moreover, the cross-seed shuffled-entropy results and the absence of a completed length-only control prevent attributing the observed modeling gain specifically to exact entropy--patch correspondence. Overall, the results identify low-dimensional patch-granularity signals as a promising basis for conditional computation in tokenizer-free LLMs. Future work should isolate entropy from correlated patch properties, improve expert utilization and sparse execution efficiency, and evaluate the approach across model scales, pretraining settings, and data domains.

\bibliography{aaai2027}

\end{document}